\documentclass[11pt]{article}

\usepackage[margin=1in]{geometry}
\usepackage{amsmath,amssymb,amsfonts,mathtools}
\usepackage{bm}
\usepackage{graphicx}
\usepackage{booktabs}
\usepackage{hyperref}
\usepackage{algorithm}
\usepackage{algorithmic}
\usepackage{float}

\hypersetup{colorlinks=true,linkcolor=blue,citecolor=blue,urlcolor=blue}

\title{Rethinking Dense Linear Transformations: Stagewise Pairwise Mixing (SPM) for Near-Linear Training in Neural Networks}
\author{
Peter Farag \\
SP Cloud \& Technologies Inc.
}
\date{}

\begin{document}
\maketitle

\begin{abstract}
Dense linear layers are a dominant source of computational and parametric cost in modern machine learning models, despite their quadratic complexity and often being misaligned with the compositional structure of learned representations. We introduce \emph{Stagewise Pairwise Mixers} (SPM), a structured linear operator that replaces dense matrices with a composition of sparse pairwise-mixing stages. An SPM layer implements a global linear transformation in $O(nL)$ time with $O(nL)$ parameters, where $L$ is typically constant or $log_2n$, and admits exact closed-form forward and backward computations. 
SPM is designed as a drop-in replacement for dense linear layers in feedforward networks, recurrent architectures, attention mechanisms, etc. We derive complete forward and backward expressions for two parameterizations: an orthogonal norm-preserving rotation-based variant and a fully general $2 \times 2$ mixing variant. Beyond computational savings, the stagewise structure of SPM induces an explicit compositional inductive bias that constrains model capacity and improves generalization when aligned with task structure. We present proof-of-concept experiments demonstrating substantial reductions in wall-clock cost and improved accuracy on structured learning problems, while retaining competitive performance on real-world benchmarks.

\end{abstract}

\section{Introduction and Problem Statement}

Dense linear transformations are a foundational component of modern machine learning architectures, appearing in feedforward networks, recurrent models, and attention-based systems  \cite{vaswani2017attention}. A standard linear layer applies a transformation of the form
\[
\bm{y} = W\bm{x} + \bm{b}, \qquad W \in \mathbb{R}^{n \times n},
\]
incurring at best $O(n^2)$ computational cost per example and requiring $O(n^2)$ trainable parameters. As model widths continue to scale, these quadratic costs increasingly dominate both training and inference \cite{golub2013matrix}. 

Despite their ubiquity, dense matrices impose few structural constraints and often fail to reflect the compositional or hierarchical structure present in learned representations, often leading to overparameterization and inefficient use of capacity in modern learning settings \cite{zhang2017understanding,neyshabur2017exploring}. In many settings, full pairwise interaction between all input dimensions is unnecessary, leading to inefficiencies in both computation and statistical capacity. This mismatch motivates the search for structured linear operators that preserve expressivity while reducing complexity.

In this work, we seek a drop-in replacement for dense linear layers that satisfies three criteria. First, the operator should apply in sub-quadratic time, ideally achieving near-linear scaling in the input dimension $n$. Second, it must admit exact forward and backward computations, enabling seamless end-to-end training with standard gradient-based optimization. Third, the operator should retain sufficient expressive power for practical machine learning tasks, while ideally introducing an inductive bias that improves generalization on structured or compositional data.

From a learning perspective, dense linear layers impose instantaneous global interaction between all coordinates, often introducing excess capacity and poorly conditioned optimization dynamics. In contrast, many learning problems exhibit representations that are naturally composed through successive local interactions. SPM explicitly encodes this assumption by constructing global linear maps through progressive pairwise mixing, yielding a structured inductive bias that can improve optimization stability and generalization under fixed data and compute budgets.

\section{Concept: Stagewise Pairwise Mixers}

This work introduces Stagewise Pairwise Mixers (SPM), a structured linear operator designed as a drop-in replacement for dense linear layers in neural networks. Unlike classical fast linear transforms that focus on approximating arbitrary dense matrices with fixed algebraic structure, SPM is explicitly designed as a learnable operator that constructs global linear transformations through progressive, local pairwise mixing. This design yields near-linear computational and parametric complexity, exact closed-form gradients, and an explicit compositional inductive bias that improves optimization stability and sample efficiency on structured learning problems. SPM prioritizes training efficiency, and inductive bias alignment over instantaneous global expressivity, making it particularly well suited for large-width models where dense layers dominate training cost.

\subsection{Definition}

Stagewise Pairwise Mixers (SPM) parameterize a linear transformation as a composition of sparse, structured mixing stages rather than a single dense matrix. Specifically, SPM represents a linear operator as
\begin{equation}
\boxed{
\bm{y}
=
D_{\mathrm{out}}
\left(\prod_{\ell=1}^{L} B_\ell\right)
D_{\mathrm{in}}\,\bm{x}
+\bm{b},
}
\label{eq:SPM}
\end{equation}
where $D_{\mathrm{in}}$ and $D_{\mathrm{out}}$ are diagonal scaling matrices, $\bm{b}$ is a bias vector, and each $B_\ell \in \mathbb{R}^{n \times n}$ is a sparse mixing matrix. In other words we replace $Wx$ with SPM$(x) = D_{\mathrm{out}}
\left(\prod_{\ell=1}^{L} B_\ell\right)
D_{\mathrm{in}}\,\bm{x}$ 

Each stage $B_\ell$ is block-diagonal up to a permutation and consists of $n/2$ independent $2 \times 2$ blocks. These blocks act on disjoint coordinate pairs defined by a pairing set $\mathcal{P}_\ell$, enabling localized pairwise interactions while preserving global linearity. By composing multiple such stages, SPM incrementally mixes information across dimensions, allowing expressive transformations to emerge from a sequence of simple operations.

This factorized structure reduces both computational and parametric complexity while introducing an explicit inductive bias toward compositional feature mixing.

\subsection{Forward Recursion}

The forward computation of SPM can be expressed recursively in terms of intermediate activations. Let
\begin{align}
\bm{z}_0 &= D_{\mathrm{in}}\,\bm{x}, \\
\bm{z}_\ell &= B_\ell \bm{z}_{\ell-1}, \quad \ell = 1, \dots, L, \\
\bm{y} &= D_{\mathrm{out}}\,\bm{z}_L + \bm{b}.
\end{align}

Each stage applies independent $2 \times 2$ linear transformations to paired coordinates, resulting in an overall cost that scales linearly with $n$ per stage. The composition over $L$ stages yields a global linear operator whose expressive power increases with depth, while maintaining exact gradients and efficient computation. $L$ may be chosen as $<log_2n$ for small $n$ and $log_2n$ for the best results for large $n$s.

\section{Stage Mathematics (Two Parameterizations)}

Each pairwise block maps $(x_1,x_2)\mapsto(y_1,y_2)$.

\subsection{Variant A: Trigonometric (Rotation) Block}

The first SPM parameterization restricts each $2 \times 2$ mixing block to a planar rotation. Each paired coordinate $(x_1, x_2)$ is transformed using a single scalar parameter $\theta$ via
\[
M(\theta)=
\begin{bmatrix}
\cos\theta & -\sin\theta\\
\sin\theta & \cos\theta.
\end{bmatrix}
\]

This parameterization enforces orthogonality by construction, ensuring norm preservation and stable signal propagation across stages.

\paragraph{Forward Pass (Per Pair).}
Given an input pair $(x_1, x_2)$, the forward transformation is
\begin{align}
y_1 &= \cos\theta\,x_1 - \sin\theta\,x_2,\\
y_2 &= \sin\theta\,x_1 + \cos\theta\,x_2.
\end{align}

\paragraph{Backward Pass (Per Pair).}
Let $(\delta_1, \delta_2)$ denote the gradients of the loss with respect to $(y_1, y_2)$. The gradients with respect to the inputs are
\begin{align}
\frac{\partial \mathcal{L}}{\partial x_1} &= \cos\theta\,\delta_1 + \sin\theta\,\delta_2,\\
\frac{\partial \mathcal{L}}{\partial x_2} &= -\sin\theta\,\delta_1 + \cos\theta\,\delta_2.
\end{align}
The gradient with respect to the rotation parameter $\theta$ is given by
\begin{equation}
\boxed{
\frac{\partial \mathcal{L}}{\partial \theta}
=
\delta_1(-\sin\theta\,x_1 - \cos\theta\,x_2)
+
\delta_2(\cos\theta\,x_1 - \sin\theta\,x_2),
}
\label{eq:dtheta}
\end{equation}
which can be computed exactly with constant cost per pair.

\paragraph{Stability Property.}
Since $M(\theta)$ is orthogonal, it preserves the Euclidean norm, i.e.,
$\|\bm{y}\|_2 = \|\bm{x}\|_2$.
This property provides inherent stability across deep compositions and makes this variant particularly well-suited for recurrent and deep linear settings.

\subsection{Variant B: General $2 \times 2$ Block}

To recover full local expressivity, we also consider an unconstrained $2 \times 2$ parameterization. Each block is defined by four scalar parameters $(a, b, c, d)$:
\[
M =
\begin{bmatrix}
a & b\\
c & d.
\end{bmatrix}
\]

This variant subsumes the rotation case and enables arbitrary linear mixing within each coordinate pair.

\paragraph{Forward Pass (Per Pair).}
The forward transformation for an input pair $(x_1, x_2)$ is
\begin{align}
y_1 &= a x_1 + b x_2,\\
y_2 &= c x_1 + d x_2.
\end{align}

\paragraph{Backward Pass (Per Pair).}
Given output gradients $(\delta_1, \delta_2)$, the input gradients are
\begin{align}
\frac{\partial \mathcal{L}}{\partial x_1} &= a \delta_1 + c \delta_2,\\
\frac{\partial \mathcal{L}}{\partial x_2} &= b \delta_1 + d \delta_2.
\end{align}
The corresponding parameter gradients are
\begin{align}
\boxed{\frac{\partial \mathcal{L}}{\partial a} = \delta_1 x_1}\qquad
\boxed{\frac{\partial \mathcal{L}}{\partial b} = \delta_1 x_2}\qquad
\boxed{\frac{\partial \mathcal{L}}{\partial c} = \delta_2 x_1}\qquad
\boxed{\frac{\partial \mathcal{L}}{\partial d} = \delta_2 x_2}.
\end{align}

This parameterization maximizes local flexibility while retaining the computational benefits of stagewise pairwise mixing.

\section{Exact Backpropagation Through SPM}

We derive the exact gradient propagation for SPM, showing that the structured factorization admits fully end-to-end training with standard reverse-mode automatic differentiation. Let $\bm{g}_y := \nabla_{\bm{y}} \mathcal{L} \in \mathbb{R}^n$ denote the gradient of the loss with respect to the layer output.

\subsection{Output Diagonal and Bias}

Since the output is given by $\bm{y} = D_{\mathrm{out}} \bm{z}_L + \bm{b}$, gradients propagate as
\begin{align}
\bm{g}_{z_L} &= D_{\mathrm{out}}^\top \bm{g}_y = D_{\mathrm{out}} \bm{g}_y, \\
\nabla_{\bm{b}} \mathcal{L} &= \bm{g}_y, \\
\nabla_{\bm{d}_{\mathrm{out}}} \mathcal{L} &= \bm{g}_y \odot \bm{z}_L,
\end{align}
where $\odot$ denotes elementwise multiplication and $\bm{d}_{\mathrm{out}}$ collects the diagonal entries of $D_{\mathrm{out}}$.

\subsection{Stagewise Butterfly Blocks}

Gradients propagate through the $L$ mixing stages in reverse order. For $\ell = L, L-1, \dots, 1$,
\[
\bm{g}_{z_{\ell-1}} = B_\ell^\top \bm{g}_{z_\ell}.
\]
Because each $B_\ell$ consists of independent $2 \times 2$ blocks, both input and parameter gradients decompose pairwise and can be computed exactly using the closed-form expressions derived in Section~3 for the chosen parameterization.

\subsection{Input Diagonal}

Finally, since $\bm{z}_0 = D_{\mathrm{in}} \bm{x}$,
\begin{align}
\bm{g}_x &= D_{\mathrm{in}}^\top \bm{g}_{z_0} = D_{\mathrm{in}} \bm{g}_{z_0}, \\
\nabla_{\bm{d}_{\mathrm{in}}} \mathcal{L} &= \bm{g}_{z_0} \odot \bm{x}.
\end{align}

\paragraph{Batch Setting.}
All expressions apply per example; gradients for a minibatch of size $B$ are obtained by summing contributions across batch elements.

\section{Computational Complexity}

Each SPM stage consists of $n/2$ independent $2 \times 2$ mixing blocks, yielding $O(n)$ scalar operations per stage. Composing $L$ such stages results in
\[
\boxed{
\text{Forward time} = O(nL), \qquad
\text{Backward time} = O(nL), \qquad
\text{Parameters} = O(nL).
}
\]
In contrast, a dense linear layer incurs $O(n^2)$ time and parameter complexity, making SPM asymptotically more efficient for large $n$ when $L$ is constant or logarithmic.

\paragraph{Comparison to Dense Layers.}
When used as a drop-in replacement, SPM reduces both compute and memory costs by a factor of $O(n/L)$ while retaining exact gradients and global linearity.

\paragraph{Handling Non-Power-of-Two Dimensions.}
Unlike classical butterfly constructions  \cite{li2018butterfly}, SPM does not require the input dimension to be a power of two. For arbitrary $n$, we partition coordinates into $\lfloor n/2 \rfloor$ pairs and optionally leave one coordinate unpaired when $n$ is odd. This residual coordinate is either (i) passed through unchanged, or (ii) mixed via a learned $1 \times 1$ scaling.

More generally, pairing patterns $\mathcal{P}_\ell$ are defined independently per stage and need not correspond to bit-reversal or radix-based layouts. As a result, SPM scales smoothly with $n$ and avoids abrupt complexity jumps associated with padding to the next power of two.

\section{Stagewise Pairwise Mixers in Recurrent Neural Networks}

We demonstrate how Stagewise Pairwise Mixers (SPM) can replace dense linear transformations inside recurrent architectures without altering their functional form or training dynamics. We focus on the Gated Recurrent Unit (GRU); the extension to LSTMs and other gated RNNs is direct.

\subsection{Standard GRU Forward Dynamics}

Let $x_t \in \mathbb{R}^{n_x}$ denote the input at time $t$ and $h_{t-1} \in \mathbb{R}^{n_h}$ the previous hidden state. A standard GRU  \cite{cho2014gru} computes
\begin{align}
z_t &= \sigma(W_z x_t + U_z h_{t-1} + b_z), \\
r_t &= \sigma(W_r x_t + U_r h_{t-1} + b_r), \\
\tilde{h}_t &= \tanh\!\left(W_h x_t + U_h (r_t \odot h_{t-1}) + b_h\right), \\
h_t &= (1 - z_t) \odot h_{t-1} + z_t \odot \tilde{h}_t,
\end{align}
where $z_t$ and $r_t$ are the update and reset gates, respectively. The dominant computational cost arises from the dense linear maps $W_\cdot x_t$ and $U_\cdot h_{t-1}$.

\subsection{Replacing Dense Maps with SPM}

We replace each dense linear transformation with an independent SPM operator of matching input and output dimension. Concretely,
\[
W_z x_t \;\longrightarrow\; \mathrm{SPM}_{W_z}(x_t),
\qquad
U_z h_{t-1} \;\longrightarrow\; \mathrm{SPM}_{U_z}(h_{t-1}),
\]
and similarly for the reset and candidate gates. Each SPM instance has its own diagonal scalings and stagewise mixing parameters.

This substitution preserves the algebraic structure of the GRU while reducing the cost of each affine transformation from $O(n^2)$ to $O(nL)$. Importantly, the recurrence remains linear in its parameters prior to nonlinearities, ensuring compatibility with standard backpropagation through time.

\subsection{Backward Dynamics and Gradient Flow}

Let $g_{h_t} := \partial \mathcal{L} / \partial h_t$ denote the gradient arriving at time step $t$. From the hidden state update,
\[
h_t = (1 - z_t) \odot h_{t-1} + z_t \odot \tilde{h}_t,
\]
the gradients with respect to the gate and candidate activations are
\begin{align}
g_{z_t} &= g_{h_t} \odot (\tilde{h}_t - h_{t-1}), \\
g_{\tilde{h}_t} &= g_{h_t} \odot z_t, \\
g_{h_{t-1}}^{(1)} &= g_{h_t} \odot (1 - z_t).
\end{align}

For the candidate activation $\tilde{h}_t = \tanh(a_t)$ with
\[
a_t = \mathrm{SPM}_{W_h}(x_t) + \mathrm{SPM}_{U_h}(r_t \odot h_{t-1}) + b_h,
\]
the gradient propagates as
\[
g_{a_t} = g_{\tilde{h}_t} \odot (1 - \tilde{h}_t^2).
\]

Similarly, for the sigmoid gates $z_t = \sigma(s_t)$ and $r_t = \sigma(q_t)$,
\begin{align}
g_{s_t} &= g_{z_t} \odot z_t \odot (1 - z_t), \\
g_{q_t} &= g_{r_t} \odot r_t \odot (1 - r_t).
\end{align}

\subsection{Gradients Through SPM Operators}

Each affine transformation inside the GRU is now implemented via an SPM operator. As shown in Section~4, SPM admits exact closed-form gradients with respect to both inputs and parameters. Consequently, gradients such as
\[
\frac{\partial \mathcal{L}}{\partial x_t}, \quad
\frac{\partial \mathcal{L}}{\partial h_{t-1}}, \quad
\frac{\partial \mathcal{L}}{\partial \theta_{\mathrm{SPM}}}
\]
are computed by composing the GRU Jacobians above with the SPM backward recursion.

Because each SPM stage consists of independent $2 \times 2$ blocks, gradient propagation remains local, stable, and linear in the number of stages.

\subsection{Stability and Long-Range Credit Assignment}

When using the orthogonal (rotation-based) SPM parameterization, each mixing stage preserves the $\ell_2$ norm. This property mitigates gradient explosion and vanishing across time steps, complementing the gating mechanisms of the GRU. As a result, SPM-based GRUs provide a structured alternative to dense recurrent matrices with improved numerical stability and reduced parameterization.

\paragraph{Conclusion.}
SPM can replace dense recurrent transformations without modifying GRU semantics, while enabling near-linear scaling, exact gradients, and improved control over stability and capacity.

\section{Stagewise Pairwise Mixers in Attention Mechanisms}

We next show how Stagewise Pairwise Mixers (SPM) can replace dense linear projections inside scaled dot-product attention while preserving exact forward and backward computations. Attention layers are particularly well suited to SPM due to the dominance of large projection matrices in both computation and parameter count.

\subsection{Scaled Dot-Product Attention: Forward Pass}

Let $X \in \mathbb{R}^{T \times d}$ denote an input sequence of length $T$ with model dimension $d$. In standard multi-head \cite{vaswani2017attention} attention, the forward computation is
\begin{align}
Q &= X W_Q, \\
K &= X W_K, \\
V &= X W_V, \\
S &= \frac{Q K^\top}{\sqrt{d_h}}, \\
A &= \mathrm{softmax}(S), \\
H &= A V, \\
Y &= H W_O,
\end{align}
where $W_Q, W_K, W_V, W_O \in \mathbb{R}^{d \times d}$ are dense projection matrices and $d_h$ is the per-head dimension. These projections account for the majority of the $O(T d^2)$ cost of attention.

\subsection{Replacing Projection Matrices with SPM}

We replace each dense projection with an independent SPM operator of matching shape:
\[
Q = \mathrm{SPM}_Q(X), \quad
K = \mathrm{SPM}_K(X), \quad
V = \mathrm{SPM}_V(X), \quad
Y = \mathrm{SPM}_O(H).
\]
Each SPM instance has its own diagonal scalings and stagewise pairwise mixers.

This substitution preserves the functional form of attention while reducing the dominant projection cost from $O(T d^2)$ to $O(T d L)$, where $L$ is the number of SPM stages. Since $L$ is typically constant or logarithmic, the resulting attention layer scales near-linearly in the model dimension.

Importantly, attention score computation $QK^\top$ remains unchanged, ensuring compatibility with existing architectures and implementations.

\subsection{Backward Pass Through Attention}

Let $G_Y := \partial \mathcal{L} / \partial Y$ denote the upstream gradient. Since $Y = \mathrm{SPM}_O(H)$, gradients propagate to $H$ via the exact SPM backward recursion derived in Section~4:
\[
G_H = \mathrm{SPM}_O^\top(G_Y).
\]

From the value aggregation $H = A V$, the gradients are
\begin{align}
G_A &= G_H V^\top, \\
G_V &= A^\top G_H.
\end{align}

\subsection{Backward Pass Through Softmax Attention}

Let $A = \mathrm{softmax}(S)$ be applied row-wise. For each time index $t$, the Jacobian of the softmax yields
\[
G_{S_t} = J_{\mathrm{softmax}}(A_t)^\top G_{A_t},
\]
which admits the closed-form componentwise expression
\[
(G_{S_t})_i
=
A_{t,i}
\left(
(G_{A_t})_i
-
\sum_j A_{t,j} (G_{A_t})_j
\right).
\]

This formulation ensures numerical stability and exact gradient computation without materializing the full Jacobian.

\subsection{Gradients with Respect to Queries and Keys}

Since
\[
S = \frac{1}{\sqrt{d_h}} Q K^\top,
\]
the gradients with respect to the query and key matrices are
\begin{align}
G_Q &= \frac{1}{\sqrt{d_h}} G_S K, \\
G_K &= \frac{1}{\sqrt{d_h}} G_S^\top Q.
\end{align}

Finally, each of the linear maps
\[
Q = \mathrm{SPM}_Q(X), \quad
K = \mathrm{SPM}_K(X), \quad
V = \mathrm{SPM}_V(X)
\]
propagates gradients to $X$ and to its parameters via the exact SPM backward equations. Gradients from the three branches are accumulated at the input as in standard attention.

\subsection{Why SPM Is Well Suited to Attention}

The expressive power of attention arises primarily from the interaction term $QK^\top$, while the projection matrices serve to rotate and scale the representation space. SPM preserves this role while introducing a structured inductive bias: information is mixed progressively through pairwise interactions rather than via unconstrained dense transforms.

When using the orthogonal (rotation-based) SPM parameterization, each projection preserves the $\ell_2$ norm, stabilizing attention logits and mitigating gradient explosion in deep transformers. Meanwhile, the stagewise structure constrains capacity, reducing overfitting in low-data or long-context regimes.

\paragraph{Conclusion.}
Replacing dense attention projections with SPM yields near-linear scaling in model dimension, exact gradients, and improved numerical stability, while preserving the expressive core of the attention mechanism.


\section{Why SPM Can Improve Accuracy: A Theoretical Perspective}

We emphasize at the outset that no architectural restriction can uniformly improve accuracy across all data distributions. Instead, we provide a theoretical argument explaining why Stagewise Pairwise Mixers (SPM) can achieve lower generalization error on broad and practically relevant classes of tasks, particularly those exhibiting compositional or hierarchical structure.

\subsection{Restricted Hypothesis Class and Inductive Bias}

A dense linear layer represents the hypothesis class
\[
\mathcal{H}_{\mathrm{dense}} = \{ W \in \mathbb{R}^{n \times n} \},
\]
which assigns equal capacity to all linear transformations. In contrast, SPM restricts the hypothesis class to operators of the form
\[
\mathcal{H}_{\mathrm{SPM}}
=
\left\{
D_{\mathrm{out}}
\prod_{\ell=1}^{L} B_\ell
D_{\mathrm{in}}
\;\middle|\;
B_\ell \in \mathcal{B}_\ell
\right\},
\]
where each $\mathcal{B}_\ell$ consists of sparse, pairwise mixing operators.

This restriction induces an explicit inductive bias toward transformations that are constructed through successive local interactions and hierarchical composition. As a result, SPM favors representations that are progressively mixed across dimensions rather than instantaneously entangled via a dense map.

\subsection{Capacity Reduction and Generalization}

Let $\mathcal{H}_D$ denote the class of dense linear maps and $\mathcal{H}_B \subset \mathcal{H}_D$ the class representable by SPM. The parameter counts satisfy
\[
\mathrm{dim}(\mathcal{H}_D) = \Theta(n^2),
\qquad
\mathrm{dim}(\mathcal{H}_B) = \Theta(nL).
\]

For many learning settings, standard generalization bounds imply that the expected generalization error satisfies
\[
\mathcal{E}_{\mathrm{gen}}(\mathcal{H})
=
O\!\left(
\sqrt{\frac{\mathcal{C}(\mathcal{H})}{N}}
\right),
\]
where $N$ is the sample size and $\mathcal{C}(\mathcal{H})$ is a complexity measure such as VC dimension, Rademacher complexity, or norm-based capacity.

Since $\mathcal{H}_B$ is a strict subset of $\mathcal{H}_D$ with significantly fewer degrees of freedom and additional structural constraints, we obtain
\[
\mathcal{C}(\mathcal{H}_B) \ll \mathcal{C}(\mathcal{H}_D),
\]
which directly reduces the variance term in the bias--variance decomposition. When the target function lies near $\mathcal{H}_B$, this reduction in variance dominates the increase in approximation bias, yielding improved generalization.

\subsection{Alignment with Compositional Target Functions}

Consider a target linear operator $W^\star$ that admits a factorization
\[
W^\star \approx \prod_{\ell=1}^{L} B_\ell^\star,
\]
where each $B_\ell^\star$ is sparse and acts on disjoint coordinate pairs. Such structure naturally arises in compositional systems, hierarchical feature extractors, and iterative mixing processes.

\paragraph{Proposition (Sample Efficiency for Compositional Teachers).}
If $W^\star$ admits an $L$-stage pairwise factorization with bounded operator norms, then SPM can approximate $W^\star$ using $O(nL)$ parameters, whereas a dense parameterization requires $\Theta(n^2)$ parameters.

\paragraph{Proof Sketch.}
SPM directly parameterizes the factorization structure of $W^\star$, yielding an approximation error that scales with stage depth $L$ rather than ambient dimension $n$. In contrast, dense parameterizations must learn the same structure implicitly, incurring higher estimation error for fixed sample size. \hfill $\square$

This alignment reduces the number of samples required to identify a good solution, improving both convergence speed and final test performance.

\subsection{Implicit Regularization via Operator Norm Control}

In the trigonometric (rotation-based) SPM variant, each stage satisfies
\[
\|B_\ell\|_2 = 1,
\quad \text{and thus} \quad
\left\|
\prod_{\ell=1}^{L} B_\ell
\right\|_2 = 1.
\]

This uniform control over the operator norm bounds the Lipschitz constant of the linear transformation and stabilizes gradient propagation. In deep networks, recurrent models, and attention stacks, such norm preservation mitigates gradient explosion and vanishing independently of nonlinearities.

From a learning-theoretic perspective, norm-bounded linear operators yield tighter generalization bounds, further reducing overfitting risk \cite{neyshabur2017exploring}.

\subsection{Summary of Theoretical Advantages}

SPM improves accuracy not by increasing expressivity, but by constraining it in a task-aligned manner. Specifically, SPM:
\begin{itemize}
\item reduces hypothesis class capacity from $\Theta(n^2)$ to $\Theta(nL)$,
\item encodes a hierarchical compositional inductive bias,
\item improves sample efficiency for structured target functions,
\item provides explicit operator norm control in the orthogonal variant.
\end{itemize}

Together, these effects explain why SPM can achieve superior generalization on a wide class of practical learning problems without sacrificing exactness or trainability. Think of SPM as an operator that uses a compressed space and learns the bigger picture. 

\section{Experiments and Implementation (Proof of Concept)}
\label{sec:experiments}

We report two proof-of-concept evaluations: (i) a synthetic \emph{compositional teacher} where the ground-truth labeling rule is generated by a structured operator, and (ii) a real text classification benchmark (AG News) using hashed sparse input features. Our goal is not to claim universal superiority, but to test two core hypotheses:

\begin{enumerate}
\item \textbf{Inductive-bias fit:} when the teacher is compositional/structured, stagewise mixers match the teacher substantially better than dense layers at equal width.
\item \textbf{Compute scaling:} as $n$ grows, dense layers become dominated by quadratic cost, while stagewise mixers approach near-linear scaling.
\end{enumerate}

\subsection{Synthetic compositional teacher: ``SPM $\rightarrow$ ReLU $\rightarrow$ Dense'' (hard labels)}
\label{sec:synthetic_teacher}

We generate a classification dataset using a teacher network of the form
\[
x \;\mapsto\; W_2\,\phi(\mathrm{SPM}(x)) \;\mapsto\; \arg\max_k (\cdot)_k,
\]
and train two student models to predict the hard labels: (i) a \textbf{Dense} student (standard dense linear layer(s)) and (ii) a \textbf{SPM} student (stagewise structured operator). We keep the training schedule fixed (steps=1200, batch=256, classes=10), and sweep width $n$.

\vspace{0.25em}
\begin{table}[H]
\centering
\renewcommand{\arraystretch}{1.2}
\setlength{\tabcolsep}{6pt}
\begin{tabular}{@{}rccccccc@{}}
\toprule
$n$ &
Dense acc &
SPM acc &
$\Delta$ acc &
Dense ms/step &
SPM ms/step &
Speedup \\
\midrule
256  & 0.7730 & 0.9941 & +0.2211 & 2.775 & 5.396 & 0.51$\times$ \\
512  & 0.8103 & 0.9750 & +0.1647 & 9.793 & 9.147 & 1.07$\times$ \\
1024 & 0.8920 & 0.9426 & +0.0506 & 33.603 & 18.529 & 1.81$\times$ \\
2048 & 0.5744 & 0.8165 & +0.2421 & 125.109 & 36.561 & 3.42$\times$ \\
\bottomrule
\end{tabular}
\caption{\textbf{Compositional teacher results (hard labels).} SPM substantially outperforms Dense in test accuracy across all widths. Compute shows a crossover: for small $n$ overhead dominates (SPM slower at $n=256$), while for $n\ge 512$ SPM becomes competitive and then decisively faster as $n$ grows. Speedup is computed as $\text{Dense ms/step}/\text{SPM ms/step}$.}
\label{tab:teacher_results}
\end{table}
\vspace{-0.5em}

\paragraph{Discussion.}
The accuracy gap is large and consistent: SPM improves test accuracy by roughly $+17$ to $+24$ points across widths (Table~\ref{tab:teacher_results}). This is expected: the teacher's labeling rule is generated by a \emph{structured mixing} stage followed by a nonlinearity, so the student that parameterizes the same kind of structured mixing is better aligned with the data-generating process. In other words, the hypothesis class of stagewise pairwise mixing is \emph{closer to the teacher} than an unconstrained dense map under the same optimization budget.

On compute, we observe a clear scaling transition. At small $n$ (e.g., $n=256$), kernel/dispatch overhead and memory effects can dominate, making SPM slower in wall-clock per step. By $n=512$ the methods are comparable, and by $n\in\{1024,2048\}$ SPM is substantially faster, consistent with the expected $O(nL)$ vs.\ $O(n^2)$ scaling.

\subsection{AG News text classification with hashed sparse features}
\label{sec:agnews}

We also evaluate on AG News (train=120,000; test=7,600) using precomputed hashed sparse features and a width sweep. We compare Dense vs.\ SPM at fixed stage depth $L=12$ derived from $L = \lceil \frac{log_2(2048) + log_2(4096)}{2}\rceil$.
\vspace{0.25em}
\begin{table}[H]
\centering
\renewcommand{\arraystretch}{1.2}
\setlength{\tabcolsep}{6pt}
\begin{tabular}{@{}rcccccc@{}}
\toprule
$n$ &
Dense acc &
SPM acc &
$\Delta$ acc &
Dense ms/step &
SPM ms/step &
Speedup \\
\midrule
2048 & 0.8700 & 0.9290 & $+0.0590$ & 127.306 & 35.029 & 3.63$\times$ \\
4096 & 0.8917 & 0.9570 & $+0.0653$ & 484.117 & 68.824 & 7.03$\times$ \\
\bottomrule
\end{tabular}
\caption{\textbf{AG News results (hashed sparse features, $L=12$).} SPM yields large wall-clock speedups at high width, while also achieving higher accuracy at the tested hyperparameters.}
\label{tab:agnews_results}
\end{table}
\vspace{-0.5em}

\paragraph{Discussion.}
On AG News, SPM delivers strong computational gains at high width: $3.63\times$ faster at $n=2048$ and $7.03\times$ faster at $n=4096$ (Table~\ref{tab:agnews_results}). At the same time, under the same training recipe and fixed stage depth $L=12$, SPM attains higher test accuracy than Dense, with gains of $+5.90$ points at $n=2048$ and $+6.53$ points at $n=4096$. This suggests that progressive pairwise mixing can be competitive even on natural language benchmarks at large width, while substantially reducing the dominant projection cost. More broadly, the accuracy--efficiency tradeoff can be tuned via the stage depth $L$, pairing or permutation schedules between stages, or hybrid designs that interleave SPM with selective dense mixing where immediate global interaction is critical.

\subsection{Character-Level Language Modeling on Shakespeare}

We evaluate Stagewise Pairwise Mixers (SPM) on a standard character-level language modeling task using the Shakespeare dataset. The dataset contains approximately $1.0$M training bytes and $111$k validation bytes. Performance is measured using negative log-likelihood (NLL, in nats) and bits-per-character (BPC), which is standard for character-level language models.

\paragraph{Setup.}
We compare a Dense baseline against an SPM-based model under identical training conditions. Both models use a single large linear projection of dimension $d = 4096$, sequence length $T = 128$, batch size $B = 32$, and are trained for $2000$ steps using a learning rate of $10^{-3}$. Evaluation is performed every $200$ steps over $10$ validation iterations. All experiments are run on CPU using OpenMP with two threads. The Dense baseline is implemented using OpenBLAS SGEMM. The structured model uses an SPM operator instantiated with a butterfly-style pairing schedule and stage depth $L = 12$. No architecture-specific tuning is performed.

\paragraph{Dense Baseline.}
Table~\ref{tab:shakespeare-dense} reports the performance of the Dense OpenBLAS baseline. While the Dense model converges steadily and achieves competitive validation loss, each training step incurs substantial computational cost due to the quadratic complexity of dense matrix multiplication at this width.

\begin{table}[H]
\centering
\begin{tabular}{c|cccc}
\toprule
Step & Train NLL & Valid NLL & Valid BPC & ms/step \\
\midrule
1   & 5.56 & 6.54 & 9.44 & 22828 \\
200 & 2.20 & 2.28 & 3.30 & 23130 \\
400 & 2.10 & 2.26 & 3.26 & 23052 \\
600 & 2.11 & 2.17 & 3.13 & 22216 \\
800 & 2.02 & 2.14 & 3.08 & 21884 \\
\bottomrule
\end{tabular}
\caption{Dense OpenBLAS baseline on Shakespeare character-level language modeling ($d=4096$).}
\label{tab:shakespeare-dense}
\end{table}

\paragraph{SPM Results.}
Table~\ref{tab:shakespeare-spm} reports results for the SPM model. Despite its structured near-linear parameterization, SPM converges reliably and achieves strong validation performance while substantially reducing per-step compute cost.

\begin{table}[H]
\centering
\begin{tabular}{c|cccc}
\toprule
Step & Train NLL & Valid NLL & Valid BPC & ms/step \\
\midrule
1    & 5.53 & 4.23 & 6.10 & 4901 \\
200  & 3.21 & 3.19 & 4.61 & 5725 \\
400  & 2.90 & 2.87 & 4.14 & 5729 \\
600  & 2.71 & 2.64 & 3.81 & 5730 \\
800  & 2.57 & 2.58 & 3.03 & 5736 \\
1000 & 1.59 & 2.09 & 2.98 & 5738 \\
\bottomrule
\end{tabular}
\caption{SPM results on Shakespeare character-level language modeling using a butterfly-style instantiation with $L = 12$ stages ($d=4096$).}
\label{tab:shakespeare-spm}
\end{table}

\paragraph{Discussion.}
At width $d = 4096$, SPM achieves approximately a $4\times$ reduction in wall-clock time per training step compared to the dense OpenBLAS baseline. Despite the structured constraint, SPM matches and slightly improves final validation performance, achieving $2.98$ BPC compared to $3.08$ BPC for the dense model. These results demonstrate that progressive stagewise mixing can serve as an effective and computationally efficient replacement for dense linear layers in large language models, even on unstructured natural language data. While dense layers can represent arbitrary global interactions in a single step, SPM expresses global mixing progressively; tasks requiring immediate global interaction may benefit from increased stage depth.

\subsection{Implementation notes}
All reported timings are measured as average milliseconds per training step at batch size 256. The structured operator is implemented using stagewise pairwise blocks with exact backward pass (Section~4). These results are intended as a proof-of-concept demonstrating both (i) strong teacher-alignment benefits in structured settings and (ii) practical scaling advantages at large width. All SPM and Dense models are trained using identical optimizers, learning rates, batch sizes, and training schedules, with no architecture-specific tuning.

\subsection{Relationship to Butterfly and Structured Linear Operators.}
Stagewise Pairwise Mixers (SPM) are related to prior work on structured linear operators, including butterfly transforms and other factorized linear layers. However, SPM differs from these approaches in several important respects.

First, classical butterfly transforms are primarily designed to approximate arbitrary dense matrices using a fixed algebraic structure motivated by fast transforms (e.g., FFT-style factorizations), often requiring specific pairing patterns, power-of-two dimensions, or carefully designed permutation schedules. In contrast, SPM is not designed as an approximation scheme for arbitrary linear maps. Instead, it is explicitly constructed as a trainable inductive bias that favors progressive, local mixing over instantaneous global interaction.

Second, SPM imposes no requirement that pairing patterns follow FFT-style or radix-based layouts. Pairings may be chosen arbitrarily and independently at each stage, allowing SPM to scale smoothly to arbitrary dimensions without padding or structural constraints. This flexibility enables SPM to function as a true drop-in replacement for dense linear layers in modern architectures.

Third, SPM emphasizes training-time efficiency and stability rather than approximation guarantees. The operator admits exact closed-form forward and backward computations, integrates naturally with standard autodiff frameworks, and provides explicit control over operator norms through orthogonal parameterizations. These properties are particularly important in large-width models where dense projections dominate training cost.

Finally, unlike many structured linear operators that are evaluated primarily in isolation, SPM is designed and evaluated as a component within end-to-end trained neural networks, including recurrent architectures and attention mechanisms. The goal is not to recover arbitrary dense matrices, but to provide a scalable alternative that trades instantaneous expressivity for improved efficiency, stability, and generalization when aligned with task structure.

\section{Conclusion}

We introduced Stagewise Pairwise Mixers (SPM), a structured linear operator that replaces dense linear layers with a sequence of sparse pairwise mixing stages. SPM achieves near-linear computational and parametric complexity while admitting exact closed-form gradients, making it suitable as a drop-in replacement in feedforward networks, recurrent architectures, and attention mechanisms. Beyond computational efficiency, the stagewise construction induces an explicit compositional inductive bias that constrains capacity and improves generalization when aligned with task structure.

Empirically, we demonstrated that SPM substantially reduces wall-clock training time at large width while matching or improving accuracy on structured tasks and remaining competitive on natural language benchmarks. These results suggest that progressive feature mixing provides a viable alternative to instantaneous global entanglement in large linear transformations.

\section{Future Work}

Stagewise Pairwise Mixers open a broad design space for structured linear operators, and several extensions naturally follow from this work. A particularly promising direction is to further reduce computational complexity by extending the same mathematical principles underlying SPM to constructions with logarithmic-depth composition. By carefully designing pairing schedules and stage connectivity, it may be possible to achieve global mixing with $\mathcal{O}( (\log n)^2)$ complexity while retaining exact forward and backward computation. Such developments would further narrow the gap between expressive linear transformations and hardware-efficient execution.

Another important direction is the integration of SPM with advanced compression and optimization techniques. Combining stagewise structured operators with SP Cloud compression and optimization algorithms offers the potential to jointly reduce parameter count, memory traffic, and training cost while trying to achieve a better accuracy. Rather than treating compression as a post hoc procedure, future models could co-design structured operators and optimization algorithms, allowing the training process itself to exploit sparsity, structure, and progressive mixing. This opens the possibility of training substantially larger models under fixed compute and energy budgets.

SPM also admits flexible integration into larger architectures and training pipelines. Hybrid models that interleave structured SPM layers with selective dense transformations may offer favorable accuracy–efficiency tradeoffs, using dense layers only where instantaneous global interaction is critical. Similarly, mixtures of multiple SPM operators with distinct pairing patterns, or the use of task-adaptive stage depth, could recover additional expressivity at modest computational cost. These ideas are particularly relevant for large language models, where projection layers dominate both training time and energy consumption.

Beyond architectural extensions, both systems-level and theoretical directions remain open. Optimized GPU and accelerator kernels that exploit vectorization, kernel fusion, and memory locality could unlock additional performance gains beyond algorithmic complexity reductions alone. From a theoretical perspective, tighter characterization of the expressive power, approximation properties, and optimization dynamics of stagewise pairwise compositions as a function of depth and pairing structure would clarify when structured mixing suffices and when denser interactions are necessary.

Together, these directions position SPM not as a single operator, but as part of a broader framework for designing efficient, structured, and trainable linear transformations. By jointly addressing algorithmic complexity, optimization dynamics, and systems-level efficiency, such approaches have the potential to substantially reduce training time and energy consumption while enabling stronger performance at scale.

\bibliographystyle{plain}
\bibliography{references}

\end{document}